\documentclass[runningheads]{llncs}
\usepackage{graphicx}
\usepackage{amsmath,amssymb} 
\usepackage{color}
\usepackage[width=122mm,left=12mm,paperwidth=146mm,height=193mm,top=12mm,paperheight=217mm]{geometry}
\usepackage{subcaption}
\usepackage{hhline}
\usepackage{dsfont}
\usepackage{color}
\usepackage{array}
\newcolumntype{L}[1]{>{\raggedright\let\newline\\\arraybackslash\hspace{0pt}}m{#1}}
\newcolumntype{C}[1]{>{\centering\let\newline\\\arraybackslash\hspace{0pt}}m{#1}}
\newcolumntype{R}[1]{>{\raggedleft\let\newline\\\arraybackslash\hspace{0pt}}m{#1}}

\begin{document}

\pagestyle{headings}
\mainmatter

\title{Towards Viewpoint Invariant 3D Human\\ Pose Estimation} 

\titlerunning{Towards Viewpoint Invariant 3D Human Pose Estimation}

\authorrunning{A. Haque, B. Peng, Z. Luo, A. Alahi, S. Yeung, and L. Fei-Fei}

\author{{\small Albert Haque, Boya Peng*, Zelun Luo*, Alexandre Alahi, Serena Yeung, Li Fei-Fei}}

\institute{Stanford University}

\normalfont

\maketitle

\begin{abstract}
We propose a viewpoint invariant model for 3D human pose estimation from a single depth image. To achieve this, our discriminative model embeds local regions into a learned viewpoint invariant feature space. Formulated as a multi-task learning problem, our model is able to selectively predict partial poses in the presence of noise and occlusion. Our approach leverages a convolutional and recurrent network architecture with a top-down error feedback mechanism to self-correct previous pose estimates in an end-to-end manner. We evaluate our model on a previously published depth dataset and a newly collected human pose dataset containing 100K annotated depth images from extreme viewpoints. Experiments show that our model achieves competitive performance on frontal views while achieving state-of-the-art performance on alternate viewpoints.
\end{abstract}

	\section{Introduction}\label{sec:intro}

	Depth sensors are becoming ubiquitous in applications ranging from security to robotics and from entertainment to smart spaces \cite{alahi2014socially}. While recent advances in pose estimation have improved performance on front and side views, most real-world settings present challenging viewpoints such as top or angled views in retail stores, hospital environments, or airport settings. These viewpoints introduce high levels of self-occlusion making human pose estimation difficult for existing algorithms.{\let\thefootnote\relax\footnote{* Indicates equal contribution.}}

	Humans are remarkably robust at predicting full rigid-body and articulated poses in these challenging scenarios. However, most work in the human pose estimation literature has addressed relatively constrained settings. There has been a long line of work on generative pose models, where a pose is estimated by constructing a skeleton using templates or priors in a top-down manner \cite{felzenszwalb2005pictorial,dantone2013human,eichner20122d,felzenszwalb2010object}. In contrast, discriminative methods directly identify individual body parts, labels, or positions and construct the skeleton in a bottom-up approach \cite{pishchulin2013poselet,pishchulin2013strong,eichner2012appearance,sapp2013modec,eichner2009better}.
	However, recent research in both classes primarily focus on frontal views with few occlusions despite the abundance of occlusion and partial-pose research in object detection \cite{rafi2015semantic,wang2013learning,azizpour2012object,ghiasi2014parsing,hsiao2014occlusion,bonde2014robust,alahi2014robust,alahi2008object,alahi2009sparsity,gao2011segmentation}.
	Even modern representation learning techniques address human pose estimation from frontal or side views \cite{li2015heterogeneous,fan2015combining,li2015maximum,tompson2014joint,jain2013learning,toshev2014deeppose,carreira2015human}. While the above methods improve human pose estimation, they fail to address viewpoint variances.

	In this work we address the problem of viewpoint invariant pose estimation from single depth images.
	There are two challenges towards this goal.
	The first challenge is designing a model that is not only rich enough to reason about 3D spatial information but also robust to viewpoint changes. The model must understand both local and global human pose structure.
	That is, it must fuse techniques from local part-based discriminative models and global skeleton-driven generative models. Additionally, it must be able to reason about 3D volumes, geometric, and viewpoint transformations.
	The second challenge is that existing real-world depth datasets are often small in size, both in terms of number of frames and number of classes \cite{ganapathi2012real,ganapathi2010real}. As a result, the use of representation learning methods and viewpoint transfer techniques has been limited.

\begin{figure}[t]
		\centering
		\includegraphics[width=1.0\linewidth]{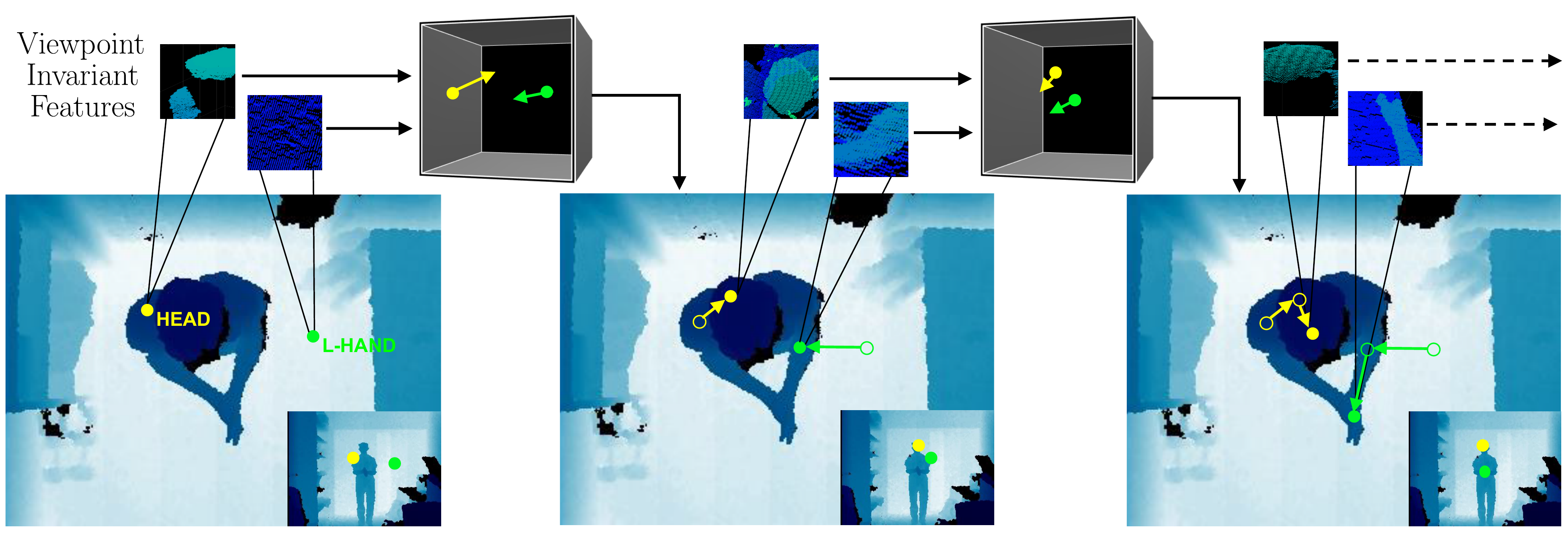}
		\caption[From a single depth image, our model uses learned viewpoint invariant feature representations to perform 3D human pose estimation with iterative refinement.]{From a single depth image, our model uses learned viewpoint invariant feature representations to perform 3D human pose estimation with iterative refinement. To provide additional three-dimensional context to the reader, a front view is shown in the lower right of each frame.}
		\label{fig:pull}
	\end{figure}

	To address these challenges, our contributions are as follows: First, on the technical side, we embed local pose information into a learned, viewpoint invariant feature space.
	Furthermore, we extend the iterative error feedback model \cite{carreira2015human} to model higher-order temporal dependencies (Figure \ref{fig:pull}).
	To handle occlusions, we formulate our model with a multi-task learning objective.
	Second, we introduce a new dataset of 100K depth images with pixel-wise body part labels and 3D human joint locations. The dataset consists of extreme cases of viewpoint variance with front, top, and side views of people performing 15 actions with occluded body parts.
	We evaluate our model on an existing public dataset \cite{ganapathi2012real} and our newly collected dataset demonstrating state-of-the-art performance on viewpoint invariant pose estimation.

	\section{Related Work}\label{sec:related}

	\textbf{RGB-Based Human Pose Estimation}. Several methods have been proposed for human pose estimation, including edge-based histograms of the human-body \cite{mori2002estimating} and silhouette contours \cite{grauman2003inferring}. More general techniques using pictorial structures \cite{felzenszwalb2005pictorial,dantone2013human,eichner20122d} and deformable part models \cite{felzenszwalb2010object}, continued to build appearance models for each local body part independently. Subsequently, higher-level part-based models were developed to capture more complex body part relationships and obtain more discriminative templates \cite{pishchulin2013poselet,pishchulin2013strong,eichner2012appearance,sapp2013modec,eichner2009better}.

	These models continued to evolve, attempting to capture even higher-level part features. Convolutional networks \cite{le1990handwritten,lecun1995convolutional}, a class of representation learning methods \cite{bengio2013representation}, began to exhibit performance gains not only in human pose estimation, but various areas of computer vision \cite{krizhevsky2012imagenet}. Since valid human poses represent a much lower-dimensional manifold in the high-dimensional input space, it is difficult to directly regress from input image to output poses with a convolutional network. As a solution to this, researchers framed the problem as a multi-task learning problem where human joints must be first detected then precisely localized \cite{li2015heterogeneous,fan2015combining,li2015maximum}.
	Jain et al. \cite{jain2013learning} enforce global pose consistency with a Markov random field representing human anatomical constraints. Follow up work by Tompson et al. \cite{tompson2014joint} combines a convolutional network part-detector with a part-based spatial model into a unified framework.

	Because human pose estimation is ultimately a structured prediction task, it is difficult for convolutional networks to correctly regress the full pose in a single pass. Recently, iterative refinement techniques have been proposed to address this issue. In \cite{sun2013deep}, Sun et al. proposed a multi-stage system of convolutional networks for predicting facial point locations. Each stage refines the output from the previous stage given a local region of the input. Building on this work, DeepPose \cite{toshev2014deeppose} uses a cascade of convolutional networks for full-body pose estimation. In another body of work, instead of predicting absolute human joint locations, Carreira et al. \cite{carreira2015human} refine pose estimates by predicting error feedback (i.e. corrections) at each iteration.

	\textbf{Depth-Based Human Pose Estimation}. Both generative and discriminative models have been proposed. Generative models (i.e. top-down approaches) fit a human body template, with parametric or non-parametric methods, to the input data. Dense point clouds provided by depth sensors motivate the use of iterative closest point algorithms \cite{ganapathi2012real,grest2005nonlinear,haehnel2003extension,knoop2006sensor} and database lookups \cite{ye2011accurate}. To further constrain the output space similar to RGB methods, graphical models \cite{he2015depth,ganapathi2010real} impose kinematic constraints to improve full-body pose estimation. Other methods such as kernel methods with kinematic chain structures \cite{ding2015articulated} and template fitting with Gaussian mixture models \cite{ye2014real} have been proposed.

	Discriminative methods (i.e. bottom-up approaches) detect instances of body parts instead of fitting a skeleton template. In \cite{shotton2011real}, Shotton et al. trained a random forest classifier for body part segmentation from a single depth image and used mean shift to estimate joint locations. This work inspired an entire line of depth-based pose estimation research exploring regression tree methods: Hough forests \cite{girshick2011efficient}, random ferns \cite{hesse2015estimating}, and random tree walks \cite{yub2015random} have been proposed in recent years.

  \textbf{Occlusion Handling and Viewpoint Invariance}.
	One popular approach to model occlusions is to treat visibility as a binary mask and jointly reason on this mask with the input images \cite{rafi2015semantic,wang2013learning}. Other approaches such as \cite{azizpour2012object,ghiasi2014parsing}, include templates for occluded versions of each part. More sophisticated models introduce occlusion priors \cite{hsiao2014occlusion,bonde2014robust} or semantic information \cite{gao2011segmentation}.

	For rigid body pose estimation and 3D object analysis, several descriptors have been proposed. Given the success of SIFT \cite{lowe1999object}, there have been several attempts at embedding rotational and translational invariance \cite{savarese20073d,wu20083d,alahi2008object}. Other features such as viewpoint invariant 3D feature maps \cite{liebelt2008viewpoint}, histograms of 3D joint locations \cite{xia2012view}, multifractal spectrum  \cite{xu2009viewpoint}, volumetric attention models \cite{haque2016recurrent}, and volumetric convolutional filters \cite{maturana20153d,maturana2015voxnet} have been proposed for 3D modeling. Instead of proposing invariant features, Ozuysal et al. \cite{ozuysal2009pose} trained a classifier for each viewpoint.
	Building on the success of representation learning from RGB, discriminative pose estimation from the depth domain, viewpoint invariant features, and occlusion modeling, we design a model which achieves viewpoint invariant 3D human pose estimation.

	\begin{figure}[t]
		\centering
		\includegraphics[width=1.0\linewidth]{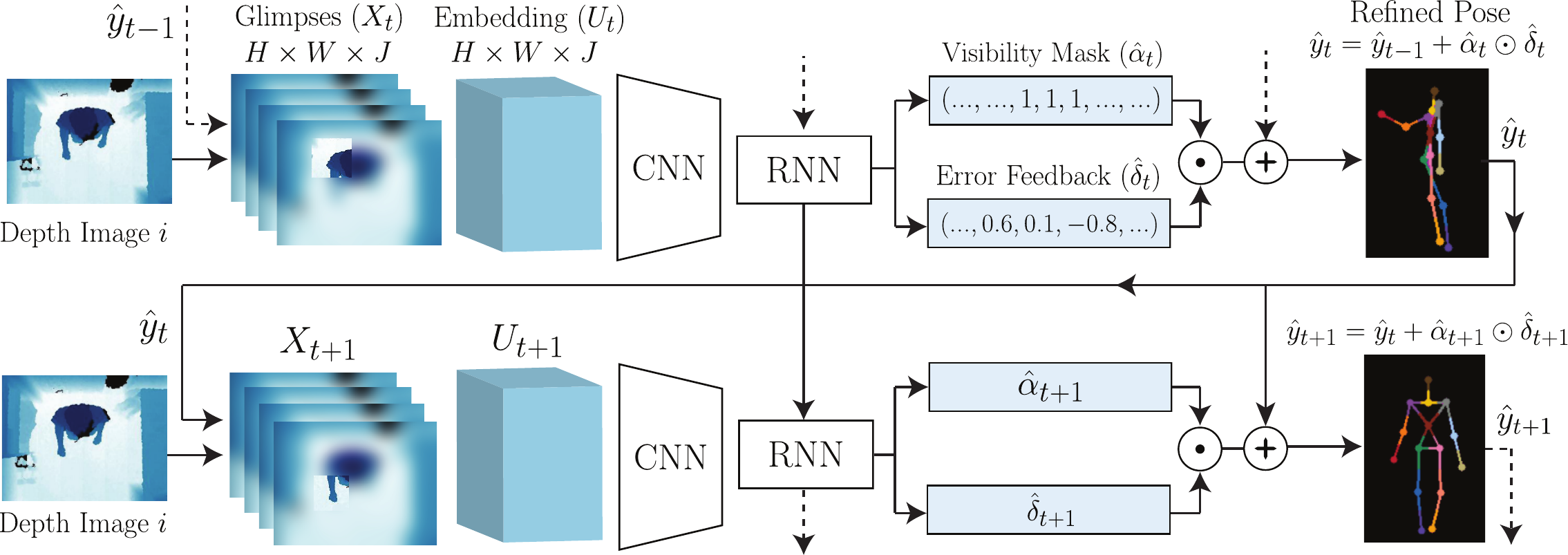}
		\caption{Model overview. The input to our model is a single depth image. We perform several iterations on this image. At iteration $t$, the input to our convolutional network is (i) a set of retina-like patches $X_t$ extracted from the input depth image and (ii) the current pose estimate $\hat{y}_{t-1}$. Our model predicts offsets $\hat{\delta}_t$ and selectively applies them to the previous pose estimate based on a predicted visibility mask $\hat{\alpha}_t$. The refined pose at the end of iteration $t$ is denoted by $\hat{y}_t$. Element-wise product is denoted by $\odot$.}
		\label{fig:model}
	\end{figure}

	\section{Model}\label{sec:model}
	\textbf{Overview.} The goal of our model is to achieve viewpoint invariant pose estimation.
	The iterative error feedback mechanism proposed by \cite{carreira2015human} demonstrates promising results on front and side view RGB images.
	However, a fundamental challenge remains unsolved: how can a model learn to be viewpoint invariant?
	Our core contribution is as follows: we leverage depth data to embed local patches into a learned viewpoint invariant feature space.
	As a result, we can train a body part detector to be invariant to viewpoint changes.
	To provide richer context, we also introduce recurrent connections to enable our model to reason on past actions and guide downstream global pose estimation (see Figure \ref{fig:model}).

	\subsection{Model Architecture}\label{sec:architecture}

	\textbf{Local Input Representation.}
	One of our goals is to use local body part context to guide downstream global pose prediction. To achieve this, we propose a two-step process. First, we extract a set of patches from the input depth image where each patch is centered around each predicted body part. By feeding these patches into our model, it can reason on low-level, local part information. We transform these patches into patches called \textit{glimpses} \cite{mnih2014recurrent,larochelle2010learning}.
	A glimpse is a retina-like encoding of the original input that encodes pixels further from the center with a progressively lower resolution.
	As a result, the model must focus on specific input regions with high resolution while maintaining some, but not all spatial information.
	These glimpses are stacked and denoted by $X \in \mathbb{R}^{H\times W \times J}$ where $J$ is the number of joints, $H$ is the glimpse height, and $W$ is the glimpse and width. Glimpses for iteration $t$ are generated using the predicted pose $\hat{y}_{t-1}$ from the previous iteration $t-1$. When $t=0$, we use the average pose $\hat{y}_0$.

	\begin{figure}[t]
		\centering
		\includegraphics[width=1.0\linewidth]{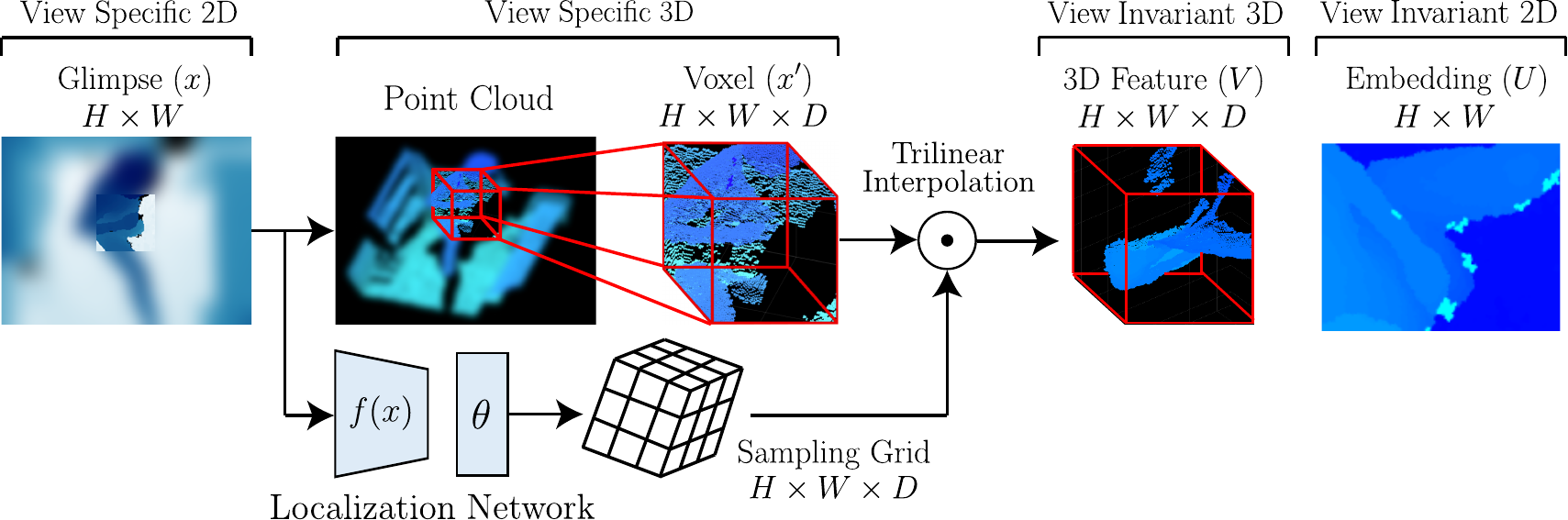}
		\caption{Learned viewpoint invariant embedding for a single glimpse. A single glimpse $x$ is converted into a voxel $x'$. A localization network $f(x)$ regresses 3D transformation parameters $\theta$ which are applied to $x'$ with a trilinear sampler. The resulting feature map $V$ is projected onto 2D which gives the embedding $U$.}
		\label{fig:stn}
	\end{figure}

	\textbf{Learned Viewpoint Invariant Embedding.} We embed the input into a learned, viewpoint invariant feature space (see Figure \ref{fig:stn}). Since each glimpse $x$ is a real world depth map, we can convert each glimpse into a voxel $x' \in \mathbb{R}^{H \times W \times D}$ where $D$ is the depth of the voxel.
	We refer to \textit{voxel} as a volumetric representation of the depth map and not a full 3D model. This representation allows us to transform the glimpse in 3D thereby simulating occlusions and geometric variations which may be present from other viewpoints.

	Given the voxel $x'$, we now transform it into a viewpoint invariant feature map $V \in \mathbb{R}^{H \times W \times D}$.
	We follow \cite{jaderberg2015spatial} in a two-step process:
    First, we use a \textit{localization network} $f(\cdot)$ to estimate a set of 3D transformation parameters $\theta$ which will be applied to the voxel $x'$.
	Second, we compute a \textit{sampling grid} defined as $G \in \mathbb{R}^{H \times W \times D}$. Each coordinate of the sampling grid, i.e. $G_{ijk} = (x^{(G)}_{ijk}, y^{(G)}_{ijk}, z^{(G)}_{ijk})$, defines where we must apply a sampling kernel in voxel $x'$ to compute $V_{ijk}$ of the output feature map.
	However, since $x^{(G)}_{ijk}, y^{(G)}_{ijk}$ and $ z^{(G)}_{ijk}$ are real-valued, we convolve $x'$ with a sampling kernel, $\ker(\cdot)$, and define the output feature map $V$:
	\begin{equation}
	\small
	V_{ijk} = \sum\limits_{a=1}^{H} \sum\limits_{b=1}^{W} \sum\limits_{c=1}^{D} x'_{abc} \ker\left( \frac{a - x^{(G)}_{ijk}}{H} \right) \ker\left(\frac{b - y^{(G)}_{ijk}}{W}\right) \ker\left(\frac{c - z^{(G)}_{ijk}}{D}\right) \label{eq:stn}
	\end{equation}
	where the kernel $\ker(\cdot) = \max(0, 1-|\cdot|)$ is the trilinear sampling kernel. As a final step, we project the viewpoint invariant 3D feature map $V$ into a viewpoint invariant 2D feature map $U$:
	\begin{equation}
	U_{ij} = \sum\limits_{c=1}^{D} V_{ijc} \textrm{\quad such that \quad} U \in \mathbb{R}^{H \times W} \label{eq:stn2}
	\end{equation}
	Notice that Equations (\ref{eq:stn}) and (\ref{eq:stn2}) are linear functions applied to the voxel $x'$. As a result, upstream gradients can flow smoothly through these mathematical units.
	The resulting $U$ now represents two-dimensional viewpoint invariant representation of the input glimpse. At this point, $U$ is used as input into a convolutional network for human body part detection and error feedback prediction.

	\textbf{Convolutional and Recurrent Networks.} As previously mentioned, our goal is to use local input patches to guide downstream global pose predictions. We stack the viewpoint invariant feature maps $U$ for each joint to form a $H \times W \times J$ tensor. This tensor is fed to a convolutional network. Through the hierarchical receptive fields of the convolutional network, the network's output is a global representation of the human pose. Directly regressing body part positions from the dense activation layers\footnote{This is referred to as \textit{direct prediction} in our experiments in Table \ref{tab:feedback}.} has proven to be difficult due to the highly non-linear mapping present in traditional human pose estimation \cite{tompson2014joint}.

	Inspired by \cite{carreira2015human}'s work in the RGB domain, we adopt an iterative refinement technique which uses multiple steps to fine-tune the pose by correcting previous pose estimates.
	In \cite{carreira2015human}, each refinement step is only indirectly influenced by previous iterations through the accumulation of error feedback.
	We claim that these refinement iterations should have a more direct and shared temporal representation.
	To remedy this, we introduce recurrent connections between each iteration; specifically a long short term memory (LSTM) module \cite{hochreiter1997long}.
	This enables our model to directly access the underlying hidden network state which generated prior feedback and model higher-order temporal dependencies.

	\subsection{Multi-Task Loss}

	Our primary goal is to achieve viewpoint invariance. In extreme cases such as top views, many human joints are occluded.
	To be robust to such occlusions, we want our model to reason on the visibility of joints.
	We formulate the optimization procedure as a multi-task problem consisting of two objectives: (i) a body-part detection task, where the goal is to determine whether a body part is visible or occluded in the input and (ii) a pose regression task, where we predict the offsets to the correct real world 3D position of visible human body joints.

	\textbf{Body-Part Detection}. For body part detection, the goal is to determine whether a particular body part is visible or occluded in the input. This is denoted by the predicted visibility mask $\hat{\alpha}$ which is a $1\times J$ binary vector, where $J$ is the total number of body joints. The ground truth visibility mask is denoted by $\alpha$. If a body part is predicted to be visible, then $\hat{\alpha}_j=1$, otherwise $\hat{\alpha}_j=0$ denotes occlusion. The visibility mask $\hat{\alpha}$ is computed using a softmax over the unnormalized log probabilities $p$ generated by the LSTM. Hence, our objective is to minimize the cross-entropy. The visibility loss for a single example is:
	\begin{equation}
	\mathcal{L}_\alpha = - \sum\limits_{j=1}^{J} \alpha_{j} \log(p_{j}) + (1-\alpha_{j}) \log(1-p_{j})  \label{eq:visibility_loss}
	\end{equation}
	Regardless of the ground truth and the predicted visibility mask, the above formulation forces our model to improve its part detection. Additionally, it allows for occluded body part recovery if the ground truth visibility is fixed to $\alpha = \mathbf{1}$.

	\textbf{Partial Error Feedback.} Ultimately, our goal is to predict the location of the joint corresponding to each visible human body part. To achieve this, we refine our previous pose prediction by learning correction offsets (i.e. feedback) denoted by $\delta$. Furthermore, we only learn correction offsets for joints that are visible. At each time step, a regression predicts offsets $\hat{\delta}$ which are used to update the current pose estimate $\hat{y}$. Specifically: $\hat{\delta}, \delta, \hat{y}, y \in \mathbb{R}^{J \times 3}$ denote real-world $(x,y,z)$ positions of each joint.
	\begin{equation}
	\mathcal{L}_\delta = \sum\limits_{j=1}^{J} \mathds{1}\{\alpha_{j} = 1\} || \hat{\delta}_j - \delta_j ||^2_2 \label{eq:error_loss}
	\end{equation}
	The loss shown in (\ref{eq:error_loss}) is motivated by our goal of predicting partial poses. Consider the case of when the right knee is not visible in the input.
	If our model successfully labels the right knee as occluded, we wish to prevent the error feedback loss from backpropagating through our network.
	To achieve this, we include the indicator term $\mathds{1}\{\alpha_{j} = 1\}$ which only backpropagates pose error feedback if a particular joint is visible in the original image. A secondary benefit is that we do not force the regressor to output dummy real values (if a joint is occluded) which may skew the model's understanding of output magnitude.

	\textbf{Global Loss.} The resulting objective is the linear combination of the error feedback cost function for all joints and the detection cost function for all body parts: $\mathcal{L} = \lambda_\alpha \mathcal{L}_\alpha + \lambda_\delta \mathcal{L}_\delta$. The mixing parameters $\lambda_\alpha$ and $\lambda_\delta$ define the relative weight of each sub-objective.

	\subsection{Training and Optimization}

	We train the full model end-to-end in a single step of optimization. We train the convolutional and recurrent network from scratch with all weights initialized from a Gaussian with $\mu=0, \sigma=0.001$. Gradients are computed using $\mathcal{L}$ and flow through the recurrent and convolutional networks. We use the Adam \cite{kingma2014adam} optimizer with an initial learning rate of $1 \times 10^{-5}$, $\beta_1=0.9$, and $\beta_2=0.999$. An exponential learning rate decay schedule is applied with a decay rate of 0.99 every 1,000 iterations.

	\section{Datasets}\label{sec:datasets}
\begin{figure}[t]
		\centering
		\begin{subfigure}[b]{0.3\linewidth}
			\includegraphics[width=\textwidth]{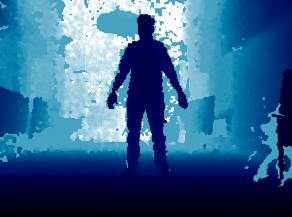}
			\caption{EVAL \cite{ganapathi2012real}}
		\end{subfigure}
		~
		\begin{subfigure}[b]{0.3\linewidth}
			\includegraphics[width=\textwidth]{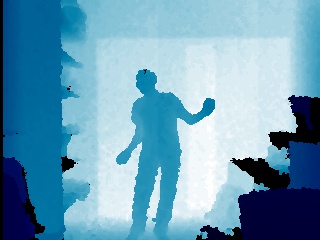}
			\caption{ITOP (Front)}
		\end{subfigure}
		~
		\begin{subfigure}[b]{0.3\linewidth}
			\includegraphics[width=\textwidth]{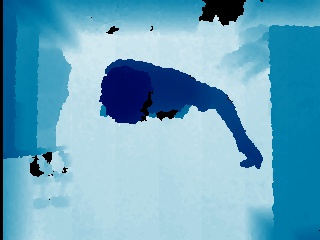}
			\caption{ITOP (Top)}
		\end{subfigure}
		\caption{Examples images from each of the datasets. Our newly collected ITOP dataset contains challenging front and top view images.}\label{fig:datasets}
	\end{figure}

	We evaluate our model on a publicly available dataset that has been used by recent state-of-the-art human pose methods. To more rigorously evaluate our model, we also collected a new dataset consisting of varied camera viewpoints. See Figure \ref{fig:datasets} for samples.

	\textbf{Previous Depth Datasets. }
	We use the Stanford EVAL dataset \cite{ganapathi2012real} which consists of 9K front-facing depth images. The dataset contains 3 people performing 8 action sequences each. The EVAL dataset was recorded using the Microsoft Kinect camera at 30 fps.  Similar to leave-one-out cross validation, we adopt a leave-one-out train-test procedure. One person is selected as the test set and the other two people are designated as the training set. This is performed three times such that each person is the test set once.

	\textbf{Invariant-Top View Dataset (ITOP).}
	Existing depth datasets for pose estimation are often small in size, both in the number of people and number of frames per person \cite{ganapathi2010real,ganapathi2012real}. To address these issues, we collected a new dataset consisting of 100K real-world depth images from multiple camera viewpoints. Named ITOP, the dataset consists of 20 people performing 15 action sequences each. Each depth image is labeled with real-world 3D joint locations from the point of view of the respective camera. The dataset consists of two ``views," namely the front/side view and the top view. The frontal view contains $360^\circ$ views of each person, although not necessarily uniformly distributed.  The top view contains images captured solely from the top (i.e. camera on the ceiling pointed down to the floor).

	\textbf{Data Collection. }
	Two Asus Xtion PRO cameras were used. One camera was placed on the ceiling facing down while another camera was from a traditional front-facing viewpoint. To annotate each frame, we used a series of steps that progressively involved more human supervision if necessary. First, 3D joints were estimated using \cite{shotton2011real} from the front-facing camera. These coordinates were then transformed into the respective world coordinate system of each camera in the system. Second, we used an iterative ground truth error correction technique based on per-pixel labeling using k-nearest neighbors and center of mass convergence. Finally, humans manually validated, corrected, and discarded noisy frames. On average, the human labeling procedure took one second per frame.

	\section{Experiments}\label{sec:experiments}

	\subsection{Evaluation Metrics}

	We evaluate our model using two metrics. As introduced in \cite{andriluka20142d}, we use the percentage of correct keypoints (PCKh) with a variable threshold.
	This metric defines a successful human joint localization if the predicted joint is within 50\% of the head segment length to the ground truth joint.

	For summary tables and figures, we use the mean average precision (mAP) which is the average precision for all human body parts. Precision is reported for individual body parts. A successful detection occurs when the predicted joint is less than 10 cm from the ground truth in 3D space.

	\subsection{Implementation Details}\label{sec:implementation_details}

	Our model is implemented in TensorFlow \cite{abaditensorflow}. We use mini-batches of size 10 and 10 refinement steps per batch. We use the VGG-16 \cite{simonyan2014very} architecture for our convolutional network but instead modify the first layer to accommodate the increased number of input channels.
	Additionally, we reduce the number of neurons in the dense layers to 2048. We remove the final softmax layer and use the second dense layer activations as input into a recurrent network. For the recurrent network, we use a long short term memory (LSTM) module \cite{hochreiter1997long} consisting of 2048 hidden units. The LSTM hidden state is duplicated and passed to a softmax layer and a regression layer for loss computation and pose-error computation. The model is trained from scratch.

	The grid generator is a convolutional network with four layers. Each layer contains: (i) a convolutional layer with 32 filters of size $3 \times 3$ with stride 1 and padding 1, (ii) a rectified linear unit \cite{nair2010rectified}, (iii), a max-pooling over a $2 \times 2$ region with stride 2. The fourth layer's output is $10\times 10 \times 32$ and is connected to a dense layer consisting of 12 output nodes which defines $\theta$. The specific 3D transformation parameters are defined in \cite{jaderberg2015spatial}.

	To generate glimpses for the first refinement iteration, the mean 3D pose from the training set is used. Glimpses are 160 pixels in height and width and centered at each joint location (in the image plane). Each glimpse consists of 4 patches where each patch is quadratically downsampled according to the patch number (i.e. its distance from the glimpse center). The input to our convolutional network is $160 \times 160 \times J$ where $J$ is the number of body part joints.

	\subsection{Comparison with State-of-the-Art}
	We compare our model to three state-of-the-art methods: random forests \cite{shotton2011real}, random tree walks (RTW) \cite{yub2015random}, and iterative error feedback (IEF) \cite{carreira2015human}. One of our primary goals is to achieve viewpoint invariance. To evaluate this, we perform three sets of experiments, progressing in level of difficulty. First, we train and test all models on front view images. This is the classical human pose estimation task. Second, we train and test all models on top view images. This is similar to the classical pose estimation task but from a different viewpoint. Third, we train on front view images and test on top view images. This is the most difficult experiment and truly tests a model's ability to learn viewpoint transfer.

	\textbf{Baselines.} We give a brief overview of the baseline algorithms:

	\noindent 1. The random forest model \cite{shotton2011real} consists of multiple decision trees that traverse each pixel to find the body part labels for that pixel. Once pixels are classified into body parts, joint positions are found with mean shift \cite{comaniciu2002mean}.

	\noindent 2. Random tree walk (RTW) \cite{yub2015random} trains a regression tree to estimate the probability distribution to the direction toward the particular joint, relative to the current position. At test time, the direction for the random walk is randomly chosen from a set of representative directions.

	\noindent 3. Iterative error feedback (IEF) \cite{carreira2015human} is a self-correcting model used to progressively make changes to an initial pose estimation by using \textit{error feedback}.

	\begin{figure}[t]
		\centering
		\begin{subfigure}[b]{0.31\textwidth}
			\includegraphics[width=\textwidth]{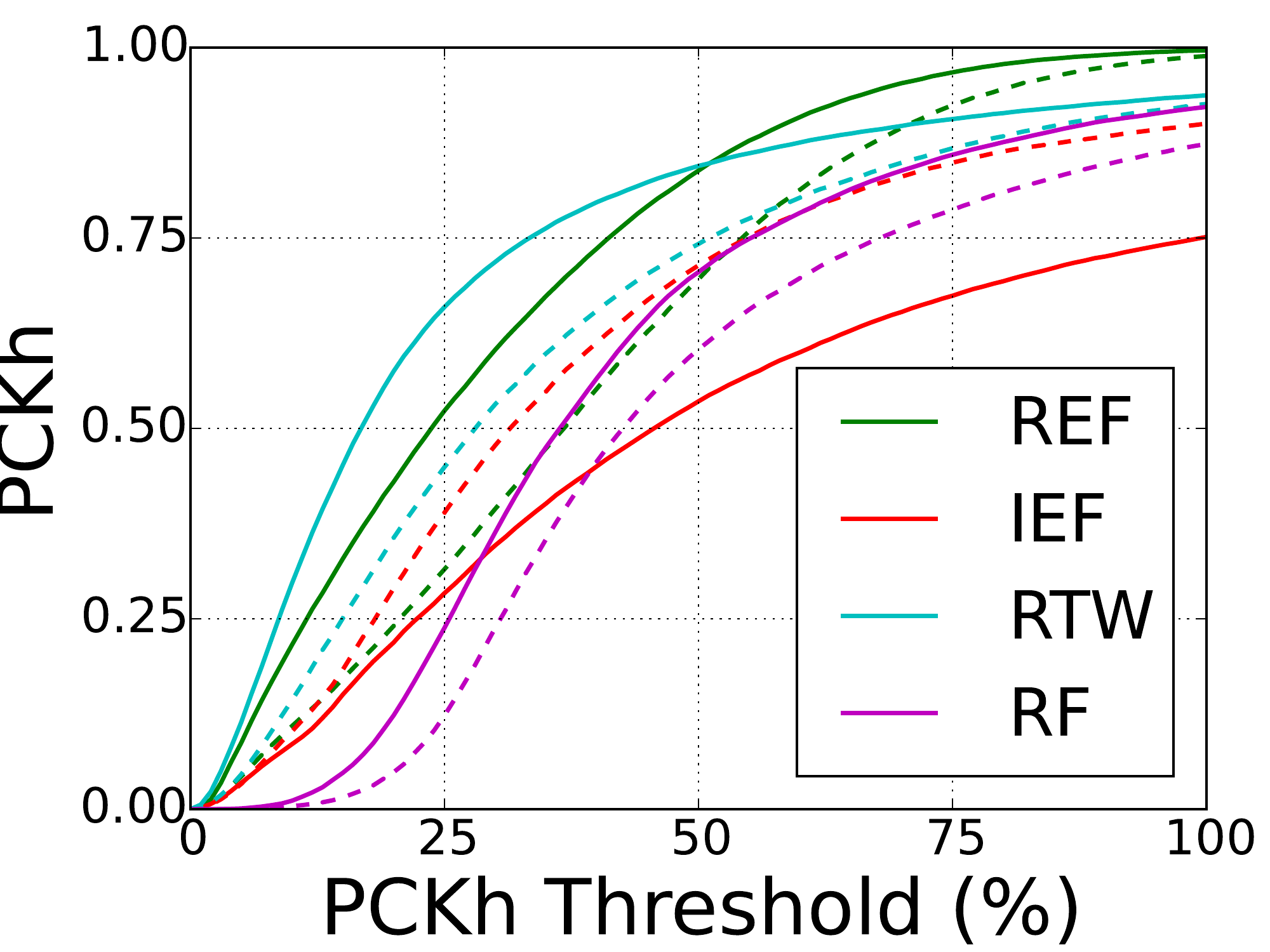}
			\caption{ITOP (front-view)}
		\end{subfigure}
		~
		\begin{subfigure}[b]{0.31\textwidth}
			\includegraphics[width=\textwidth]{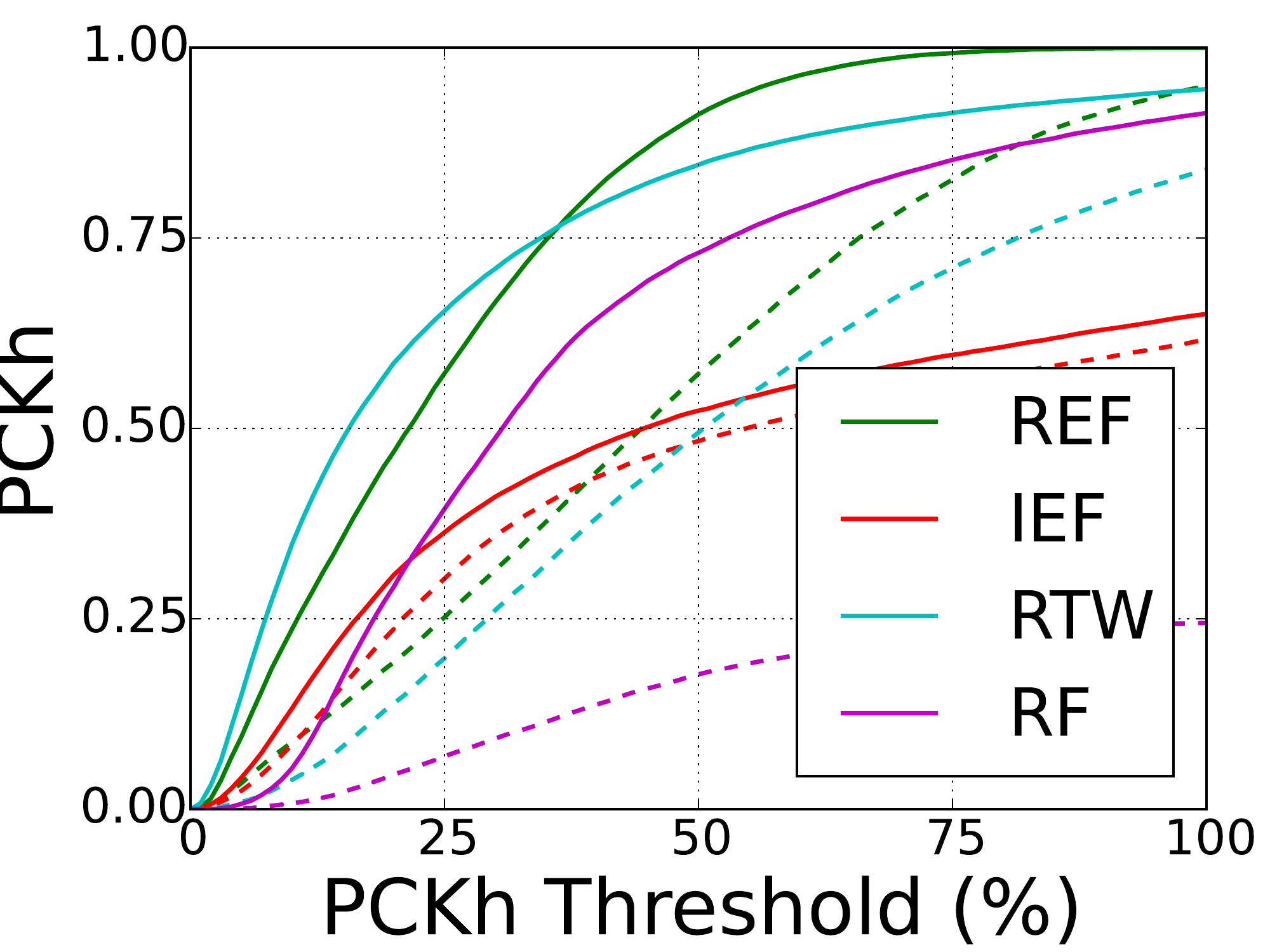}
			\caption{ITOP (top-view)}
		\end{subfigure}
		~
		\begin{subfigure}[b]{0.31\textwidth}
			\includegraphics[width=\textwidth]{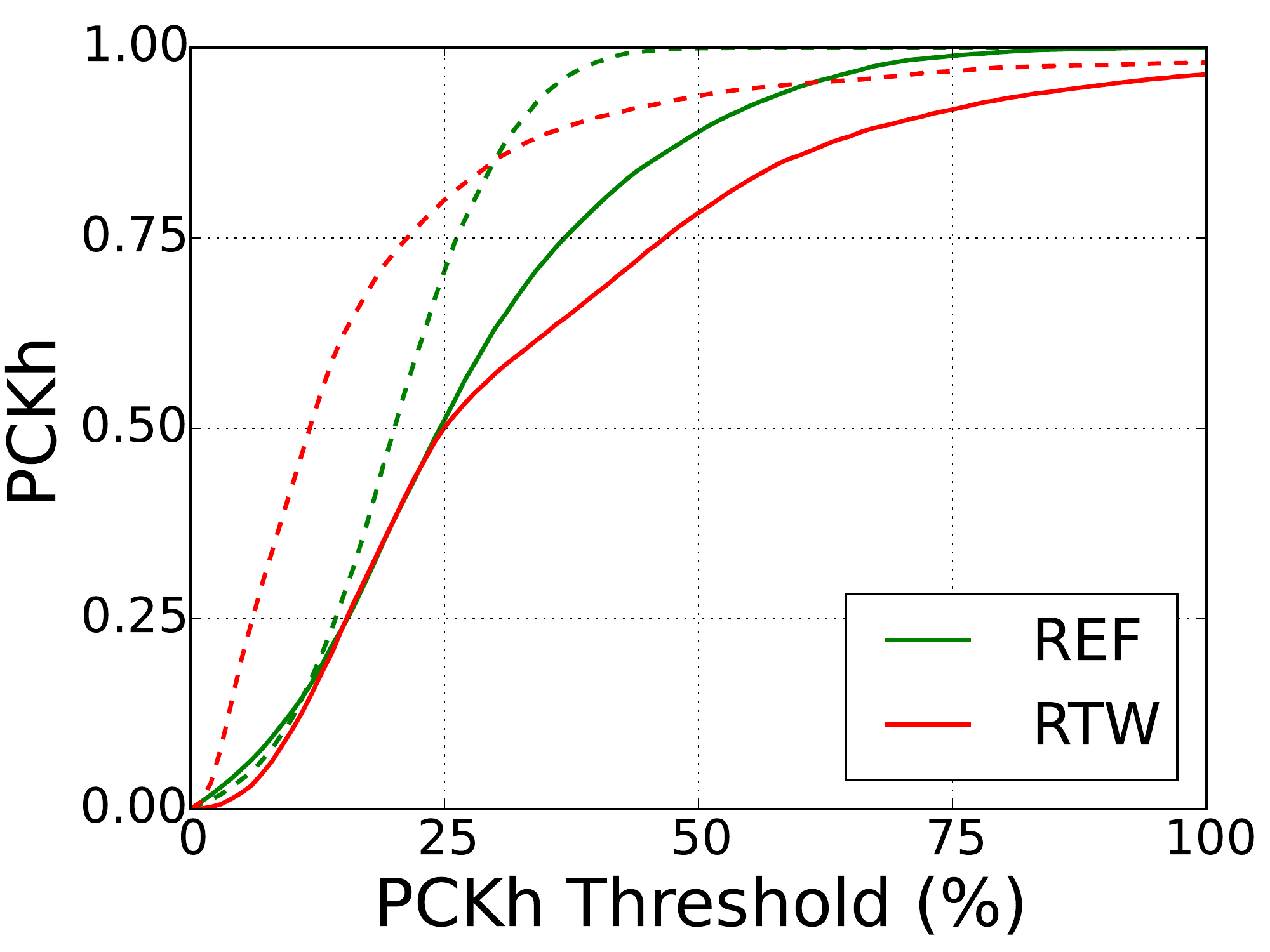}
			\caption{EVAL}
		\end{subfigure}
		\caption{Percentage of correct keypoints based on the head (PCKh). Colors indicate different methods. Solid lines indicate full body performance. Dashed lines indicate upper body performance. Higher is better.}
		\label{fig:pckh}
	\end{figure}

	\begin{table}[t]
		\centering
		\begin{tabular}{l|C{9mm}C{9mm}C{9mm}C{9mm}|C{9mm}C{9mm}C{9mm}C{9mm}|C{9mm}C{9mm}}
			\hline
			& \multicolumn{4}{c|}{ITOP (front-view)} & \multicolumn{4}{c|}{ITOP (top-view)} & \multicolumn{2}{c}{EVAL}\\ \hline
Body Part & RTW & RF & IEF & Ours & RTW & RF & IEF & Ours & RTW  & Ours \\ \hline
Head & 97.8 & 63.8 & 96.2 & 98.1 & 98.4 & 95.4 & 83.8 & 98.1 & 90.9 & 93.9 \\
Neck & 95.8 & 86.4 & 85.2 & 97.5 & 82.2 & 98.5 & 50.0 & 97.6 & 87.4 & 94.7 \\
Shoulders & 94.1 & 83.3 & 77.2 & 96.5 & 91.8 & 89.0 & 67.3 & 96.1 & 87.8 & 87.0 \\
Elbows & 77.9 & 73.2 & 45.4 & 73.3 & 80.1 & 57.4 & 40.2 & 86.2 & 27.5 & 45.5 \\
Hands & 70.5 & 51.3 & 30.9 & 68.7 & 76.9 & 49.1 & 39.0 & 85.5 & 32.3 & 39.6 \\ \hline
Torso & 93.8 & 65.0 & 84.7 & 85.6 & 68.2 & 80.5 & 30.5 & 72.9 & --- & --- \\
Hips & 80.3 & 50.8 & 83.5 & 72.0 & 55.7 & 20.0 & 38.9 & 61.2 & --- & --- \\
Knees & 68.8 & 65.7 & 81.8 & 69.0 & 53.9 & 2.6 & 54.0 & 51.6 & 83.4 & 86.0 \\
Feet & 68.4 & 61.3 & 80.9 & 60.8 & 28.7 & 0.0 & 62.4 & 51.5 & 90.0 & 92.3 \\
\hhline{===========}
Upper Body & 84.8 & 70.7 & 61.0 & 84.0 & 84.8 & 73.1 & 51.7 & 91.4 & 59.2 & 73.8 \\
Lower Body & 72.5 & 59.3 & 82.1 & 67.3 & 46.1 & 7.5 & 53.3 & 54.7 & 86.7 & 89.2 \\
Full Body & 80.5 & 65.8 & 71.0 & 77.4 & 68.2 & 47.4 & 51.2 & 75.5 & 68.3 & 74.1 \\

\hline
		\end{tabular}
		\caption{Detection rates of body parts using a 10 cm threshold. Higher is better. Results for the left and right body part were averaged. Upper body consists of the head, neck, shoulders, elbows, and hands.}
		\label{table:map}
	\end{table}
	\textbf{Train on front views, test on front views.}
	Table \ref{table:map} shows the average precision for each joint using a 10 cm threshold and the overall mean average precision (mAP) while Figure \ref{fig:pckh} shows the PCKh for all models.
	IEF and the random forest methods were not evaluated on the EVAL dataset. Random forest depends on a per-pixel body part labeling, which is not provided by EVAL. IEF was unable to converge to comparable results on the EVAL dataset. We discuss the ITOP results below.
	For frontal views, RTW achieves a mAP of 84.8  and 80.5 for the upper and full body, respectively. Our recurrent error feedback (REF) model performs similarly to RTW, achieving a mAP of 2 to 3 points less.
	The random forest algorithm achieves the lowest full body mAP of 65.8. This could be attributed to the limited amount of training data. The original algorithm \cite{shotton2011real} was trained on 900K synthetic depth images.

\begin{figure}[t]
		\centering
		\hspace{-10mm}
		\includegraphics[width=0.95\linewidth]{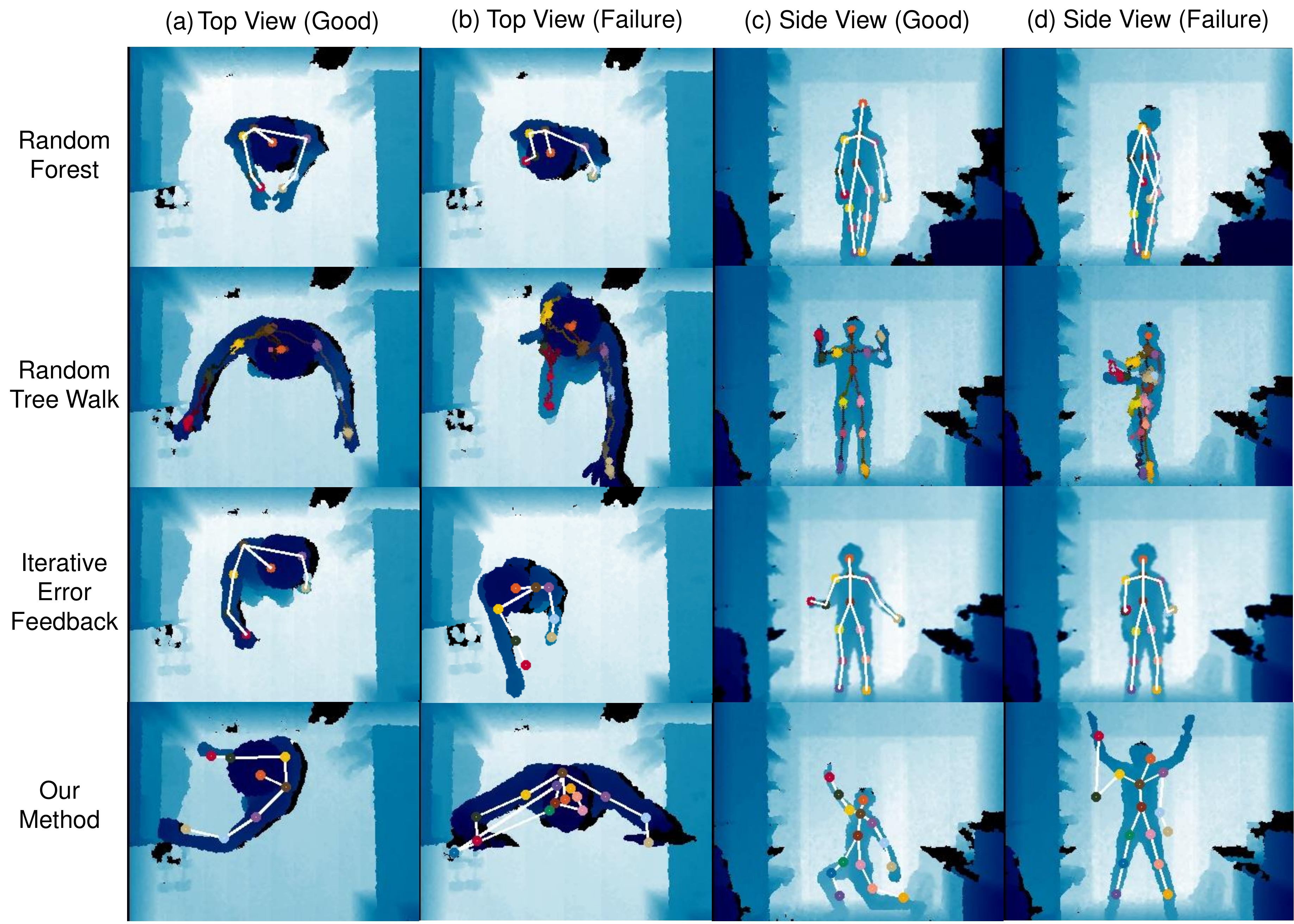}
		\caption{Qualitative results without viewpoint transfer}
		\label{fig:qualitative}
	\end{figure}

	We show qualitative results in Figure \ref{fig:qualitative}. The front-view ITOP dataset is shown in columns (c) and (d). Both our model and IEF make similar mistakes: both models sometimes fail to learn sufficient feedback to converge to the correct body part location. Since we do not impose joint position constraints or enforce skeleton priors, our method incorrectly predicts the elbow location.

	\textbf{Train on top view, test on top view.}
	Figure \ref{fig:qualitative} shows examples of qualitative results from frontal and top down views for Shotton et al. \cite{shotton2011real} and random tree walk (RTW) \cite{yub2015random}. For the top-down view, we show only 8 joints on the upper body (i.e. head, neck, left shoulder, right shoulder, left elbow, right elbow, left hand, and right hand) as the lower body joints are almost always occluded. RF and RTW give reasonable results when all joints are visible (see Figure \ref{fig:qualitative}a and \ref{fig:qualitative}c) but do not perform well in the case of occlusion (Figure \ref{fig:qualitative}b and \ref{fig:qualitative}d). For the random forest method, we can see from figure \ref{fig:qualitative}b that the prediction for the occluded right elbow is topologically invalid though both right shoulder and hand are visible and correctly predicted. This is because the model doesn't take into account the topological information among joints, so it is not able to modify its prediction for one joint base on the predicted positions of neighboring joints. For RTW, figure \ref{fig:qualitative}b shows that the predicted position for right hand goes to the right leg. Though legs and hands possess very different depth information, the model mistook the right leg for right hand when the hand is occluded and the leg appears in the common spatial location of a hand.

	\begin{table}[t]
		\centering
		\begin{tabular}{lC{1cm}C{1cm}C{1cm}C{2cm}}
			\hline
			Body Part   & RTW   & RF   & IEF  & Our Model \\ \hline
			Head        & 1.5 &	48.1 &	47.9 &	55.6 \\
			Neck        & 8.1 &	5.9	 &  39.0 &	40.9 \\
			Torso       & 3.9 &	4.7	 &  41.9 &	35.0 \\  \hline
			Upper Body  & 2.2 &	19.7 &	23.9 &	29.4 \\
			Full Body   & 2.0 &	10.8 &	17.4 &	20.4 \\ \hline
		\end{tabular}
		\caption{Detection rate for the viewpoint transfer task}
		\label{tab:transfer}
	\end{table}

	\textbf{Train on frontal views, test on top views.} This is the most difficult task for 3D pose estimation algorithms since the test set contains significant scale and shape differences from the training data. Results are shown in Table \ref{tab:transfer}. RTW gives the lowest performance as the model relies heavily on topological information.
	If the prediction for an initial joint fails, error will accumulate onto subsequent joints.
	Both deep learning methods are able to localize joints despite the viewpoint change.
	IEF achieves a 47.9 detection rate for the head while our model achieves a 55.6 detection rate.
	This can be attributed to the proximity of upper body joints in both viewpoints. The head, neck, and torso locations are similarly positioned across viewpoints.

	\textbf{Runtime Analysis.} Methods which employ deep learning techniques often require more computation for forward propagation compared to non deep learning approaches. Our model requires 1.7 seconds per frame (10 iterations, forward-pass only) while the random tree walk requires 0.1 second per frame. While this is dependent on implementation details, it does illustrate the tradeoff between speed and performance.

	\subsection{Ablation Studies}

	To further gauge the effectiveness of our model, we analyze each component of our model and provide both quantitative and qualitative analyses. Specifically, we evaluate the effect of error feedback and discuss the relevance of the input glimpse representation.

	\textbf{Effect of Recurrent Connections.}
	We analyze the effect of recurrent connections compared to regular iterative error feedback and direct prediction. To evaluate iterative feedback, we use our final model but remove the LSTM module and regress the visibility mask $\hat{\alpha}$ and error feedback $\hat{\delta}$ using the dense layer activations. Note that we still use a multi-task loss and glimpse inputs. Direct prediction does not involve feedback but instead attempts to directly regress correct pose locations in a single pass.

	\begin{table}[t]
		\centering
		\begin{tabular}{L{2cm}|C{1.5cm}C{1.5cm}|C{1.5cm}C{1.5cm}|C{1.5cm}C{1.5cm}}
			\hline
			& \multicolumn{2}{C{3cm}|}{Direct Prediction} & \multicolumn{2}{C{3cm}|}{Iterative Feedback} & \multicolumn{2}{C{3cm}}{Recurrent Feedback} \\ \hline
			Body Part & Front & Top & Front & Top & Front & Top \\ \hline
			Head & 27.8 & 32.1 & 96.2 & 83.8 & 98.1 & 98.1 \\
			Hands & 1.3 & 1.8 & 30.9 & 39.0 & 68.7 &  85.5 \\
			Upper Body & 15.0 & 17.8 & 61.0 & 51.7 & 84.0 & 91.4\\
			Full Body & 21.8 & 23.8 & 71.0 & 51.2 & 77.4 & 75.5 \\ \hline
		\end{tabular}
		\caption{Detection rate of our model with different feedback mechanisms on the ITOP front dataset. Rows denote a different body parts. Model is trained without viewpoint transfer and the detection threshold is 10 cm.}
		\label{tab:feedback}
	\end{table}
	\begin{figure}[t]
		\centering
		\includegraphics[width=1.0\textwidth]{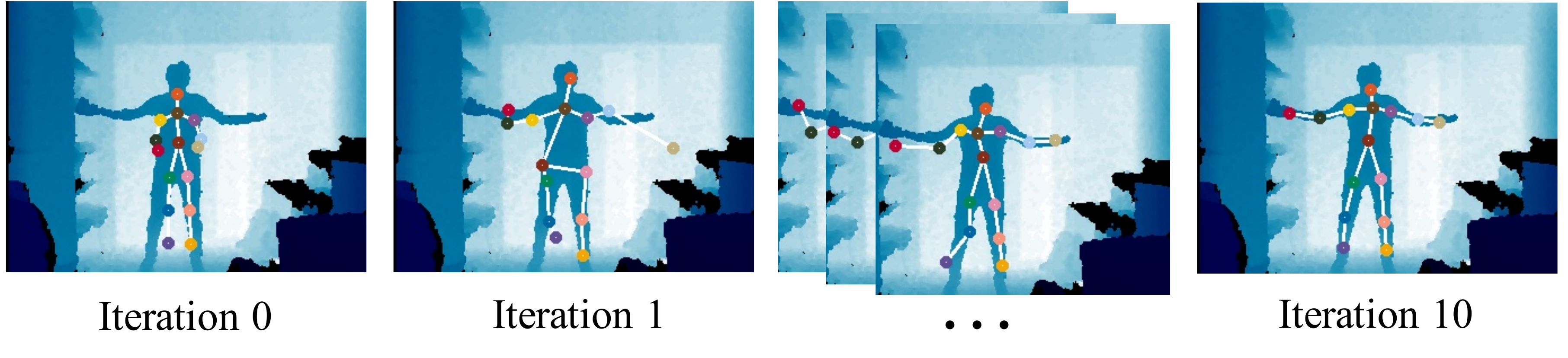}
		\caption{Our model's estimated pose at different iterations of the refinement process. Initialized with the average pose, it converges to the correct pose over time.}
		\label{fig:iterations}
	\end{figure}

	Quantitative results are shown in Table \ref{tab:feedback}. Direct prediction, as expected, performs poorly as it is very difficult to regress exact 3D joint locations in a single pass. Iterative-based approaches significantly improve performance by 30 points. It is clear that recurrent connections improve performance, especially in the top-view case where recurrent feedback achieves 91.4 upper body mAP while iterative feedback achieves 51.7 upper body mAP.

\begin{figure}[t]
		\centering
		\begin{subfigure}[b]{0.59\textwidth}
			\includegraphics[width=\textwidth]{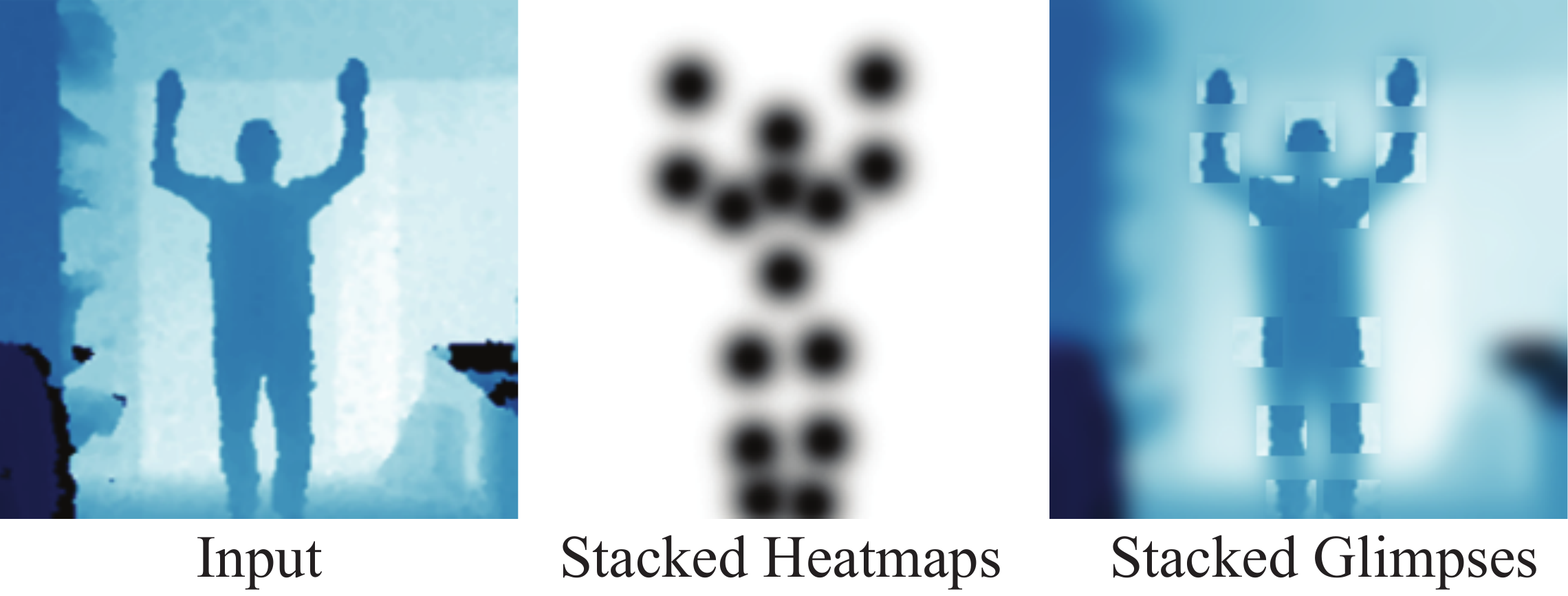}
			\caption{Heatmap vs glimpse input representation}
		\end{subfigure}
		~
		\begin{subfigure}[b]{0.38\textwidth}
			\includegraphics[width=0.9\textwidth]{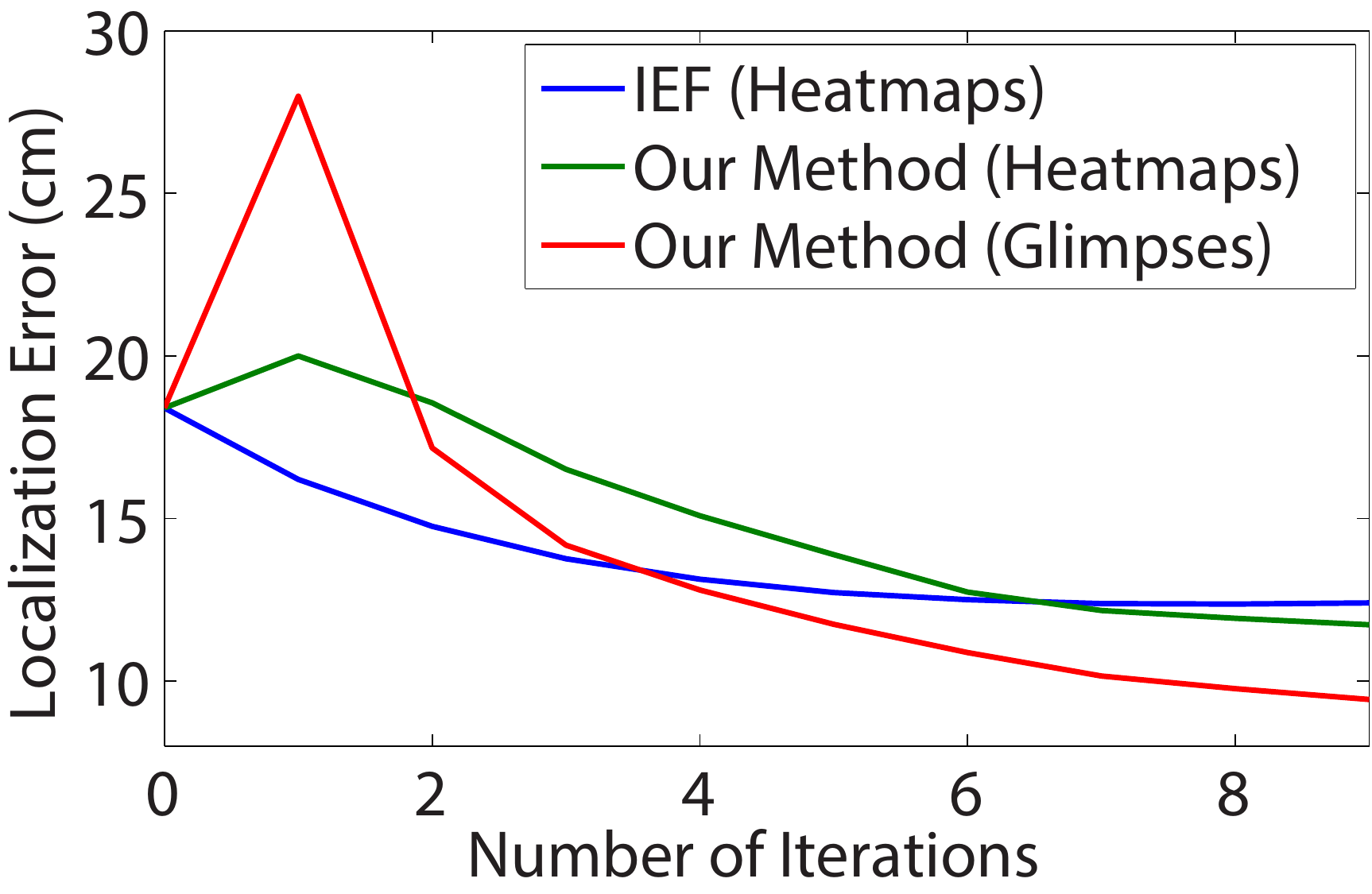}
			\caption{Localization error}
		\end{subfigure}
		\caption{Comparison of heatmap and glimpse input representations. (a) Multi-channel heatmap and glimpse input projected onto a 2D image. (b) Localization error as a function of refinement iterations. Lower error is better.}\label{fig:glimpse_vs_heatmap}
	\end{figure}

	Figure \ref{fig:iterations} shows how our model updates the pose over time. Consistent across all images, the first iteration always involves a large, seemingly random transformation of the pose. This can be thought of as the model is ``looking around" the initial pose estimate. Once the model understands the initial surrounding area, it returns to the human body and begins to fine-tune the pose prediction, as shown in iteration 10. Figure \ref{fig:glimpse_vs_heatmap}b quantitatively illustrates this result.

	\textbf{Effect of Glimpses.}
	Our motivation for glimpses is to provide additional local context to our model to guide downstream, global pose estimation. In Figure \ref{fig:glimpse_vs_heatmap} we evaluate the performance of glimpses vs indicator masks (i.e. heatmaps). Figure \ref{fig:glimpse_vs_heatmap}b shows that glimpses do provide more context for the global pose prediction task. As the number of refinement iterations increases, using glimpses, the localization error for each joint is less than the error with heatmaps. By looking at Figure \ref{fig:glimpse_vs_heatmap}a, it becomes apparent that heatmaps provide limited spatial information. The indicator mask is a way of encoding two-dimensional body part coordinates but does not explicitly provide local context information. Glimpses are able to provide such context from the input image.

	\section{Conclusion}
	We introduced a viewpoint invariant model that estimates 3D human pose from a single depth image. Our model is formulated as a deep discriminative model that attends to glimpses in the input. Using a multi-task optimization objective, our model is able to selectively predict partial poses by using a predicted visibility mask. This enables our model to iteratively improve its pose estimates by predicting occlusion and human joint offsets. We showed that our model achieves competitive performance on an existing depth-based pose estimation dataset and achieves state-of-the-art performance on a newly collected dataset containing 100K annotated depth images from several view points.

	\subsubsection{Acknowledgements.}
	We gratefully acknowledge the Clinical Excellence Research Center (CERC) at Stanford Medicine and thank the Office of Naval Research, Multidisciplinary University Research Initiatives Program (ONR MURI) for their support.

\clearpage

	\newpage
	\appendix
	\section*{Appendices}
	\section{Localization Heatmaps}
	\vspace{-2mm}
	To further analyze the viewpoint transfer task (train on front and side views, test on top views), we visualize the localization heatmap in the figures below. For each body part, we plot the predicted test-set locations with respect to the ground truth. Clusters closer to $(0,0)$ are better. All axes denote centimeters.
	
\begin{figure}[h]
		\centering
		\includegraphics[width=\textwidth]{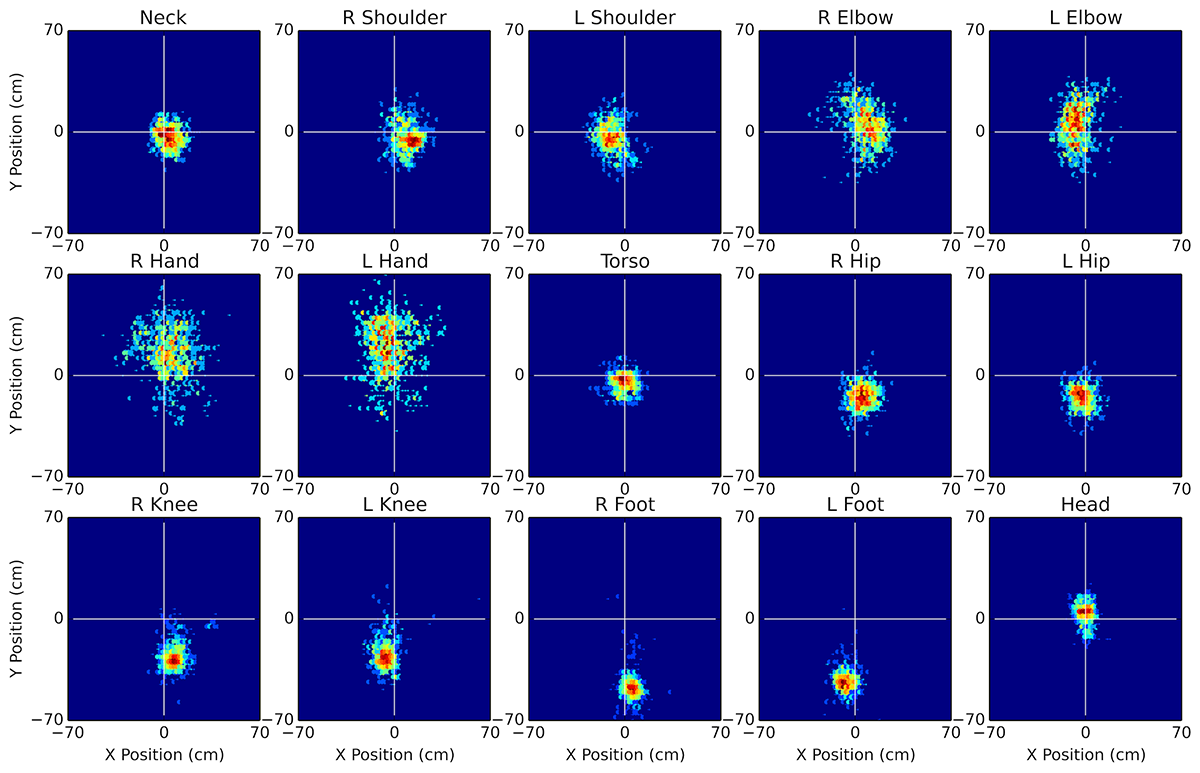}
		\vspace{-6mm}
		\caption{Predicted joint locations for our method (iteration 10) for the viewpoint transfer task. The point (0,0) indicates the ground truth location.}
		\label{fig:ref_9}
	\end{figure}

	Figure \ref{fig:ref_9} shows our model's outputs for the viewpoint transfer task. For lower body parts, our model makes a systemic error of predicting joints to be lower (i.e. closer to the ground) than the ground truth. From the top view, the lower body parts are not only further from the camera but they are also often occluded which forces our model to reason based on global pose structure as opposed to fine-tuned local information. For the upper body, most joints are visible which lead to more correct predictions.
	
	\vspace{-3mm}
	
\begin{figure}[h!]
		\centering
		\includegraphics[width=\textwidth]{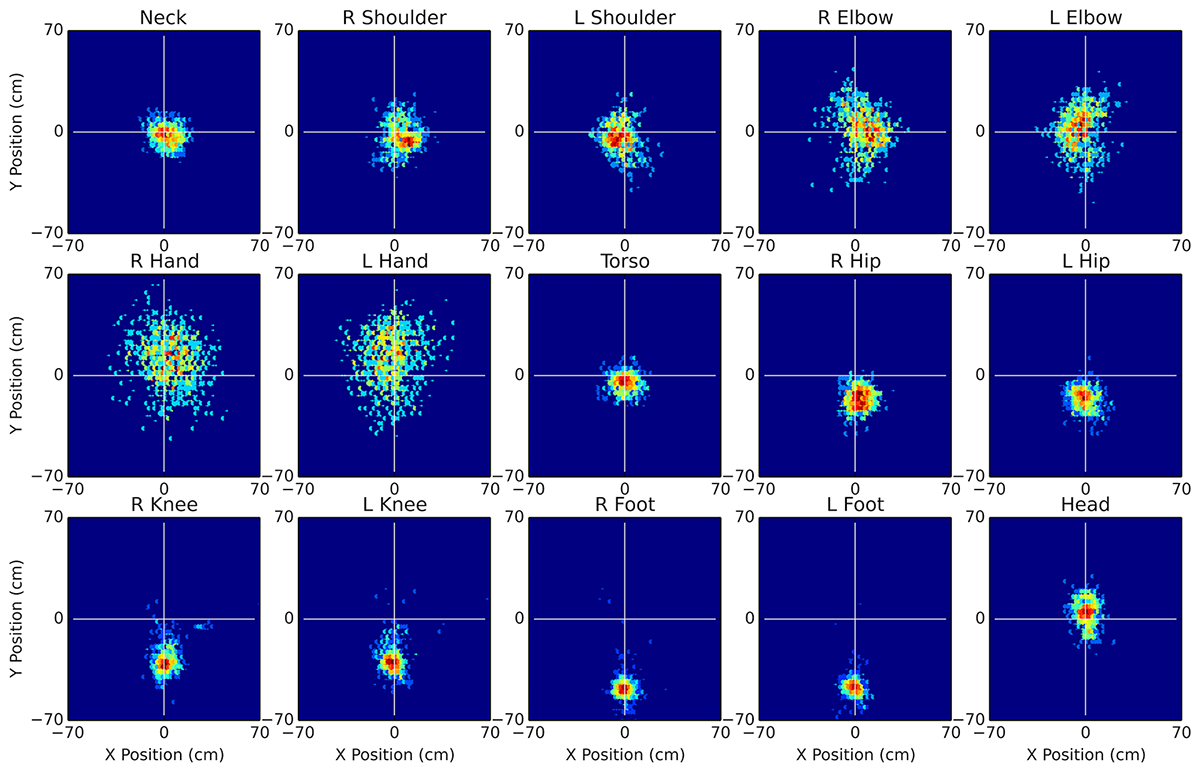}
		\vspace{-6mm}
		\caption{Predicted joint locations for iterative error feedback (iteration 0) for the viewpoint transfer task. The point (0,0) indicates the ground truth location.}\label{fig:ief_0}
		\vspace{2mm}
		\includegraphics[width=\textwidth]{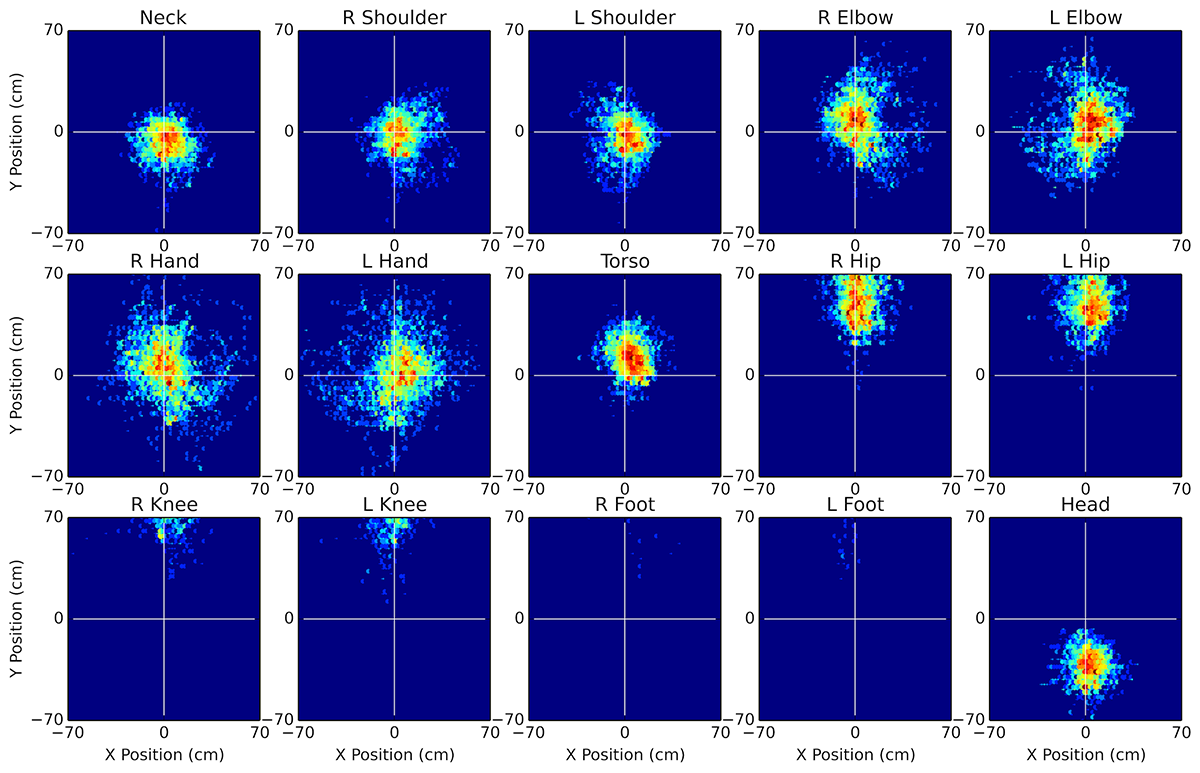}
		\vspace{-6mm}
		\caption{Predicted joint locations for our method (iteration 0) for the viewpoint transfer task. The point (0,0) indicates the ground truth location.}\label{fig:ref_0}
	\end{figure}

	\newpage
	\noindent Below, Figures \ref{fig:ief_0} and \ref{fig:ref_0} show the differences between the initialization strategies of IEF and our method.
	\vspace{-3mm}
	\vspace{-3mm}

	\newpage
	\noindent Random tree walk tends to perform poorly on the viewpoint transfer task. The heatmaps below show predictions very far from the ground truth.
	\vspace{-3mm}
	\begin{figure}[h!]
		\centering
		\includegraphics[width=\textwidth]{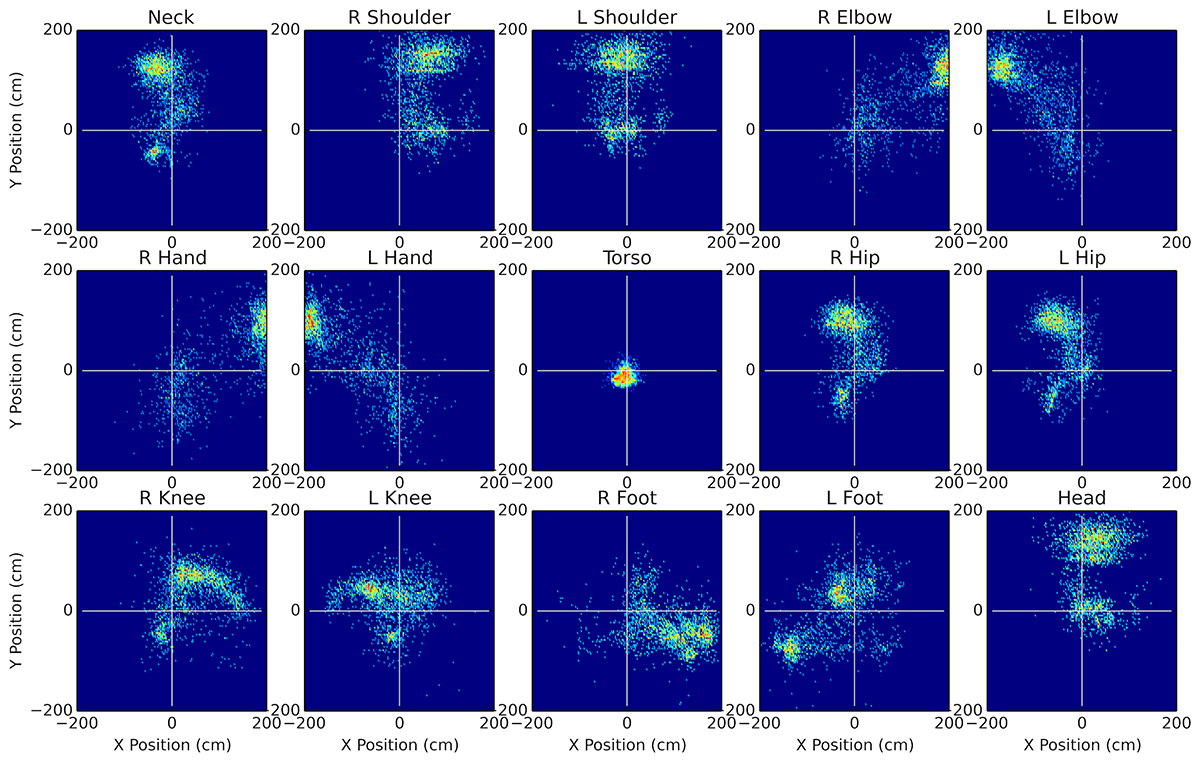}
		\vspace{-6mm}
		\caption{Predicted joint locations for random tree walk (step 0) for the viewpoint transfer task. The point (0,0) indicates the ground truth location.}\label{fig:rtw_0}
		\vspace{2mm}
		\includegraphics[width=\textwidth]{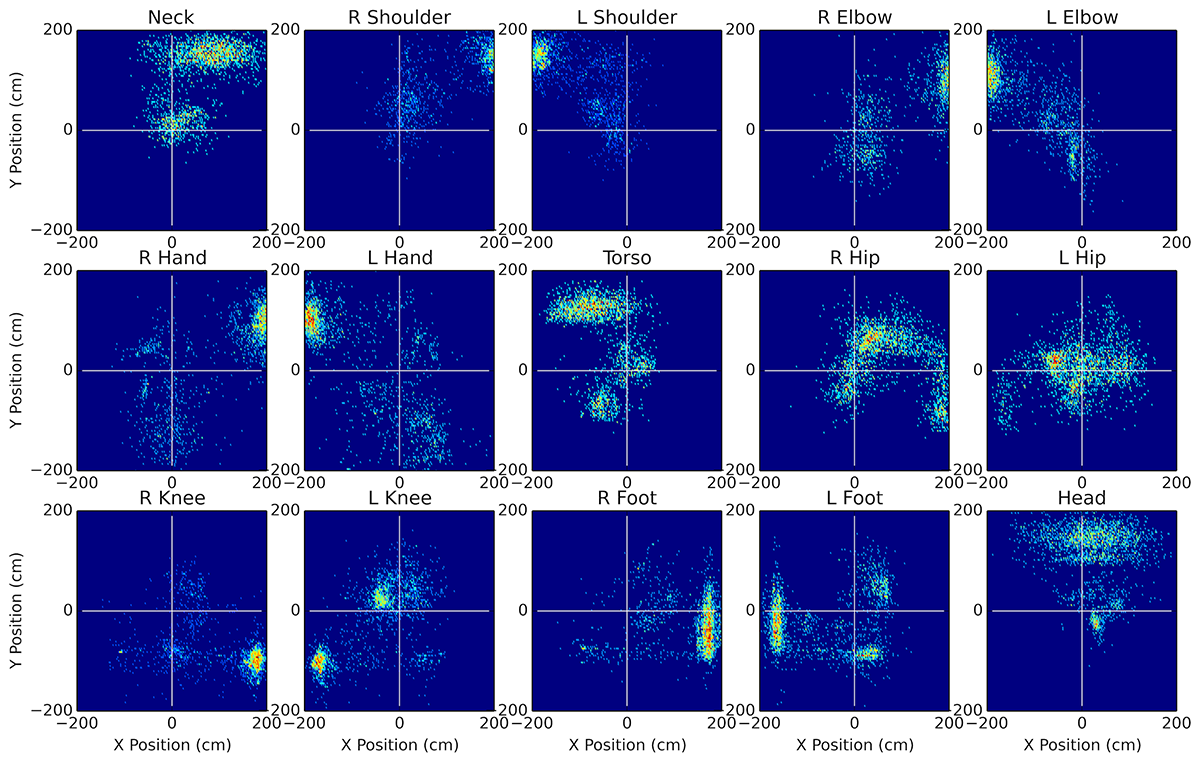}
		\vspace{-6mm}
		\caption{Predicted joint locations for random tree walk (step 300) for the viewpoint transfer task. The point (0,0) indicates the ground truth location.}
	\end{figure}
	\vspace{-3mm}

\end{document}